\documentclass[10pt,conference]{IEEEtran}
\usepackage{cite}
\usepackage{amsmath,amssymb,amsfonts}
\usepackage{algorithm}
\usepackage[algo2e]{algorithm2e}
\usepackage{graphicx}
\usepackage{float}

\newlength{\leftbarwidth}
\usepackage{amsmath,amssymb,amsfonts}
\usepackage{graphicx}
\usepackage{epstopdf}
\usepackage{balance}
\usepackage[framemethod=tikz]{mdframed}
\usepackage{lipsum}
\usepackage{soul}
\definecolor{light-gray}{gray}{0.80}
\usepackage{textcomp}
\usepackage{algorithm,algpseudocode}
\renewenvironment{quote}%
{\begin{leftbar}\noindent\hspace{-\leftbarwidth}\xspace}%
{\end{leftbar}}%
\usepackage{amsmath}

\usepackage{hyperref}
\usepackage{cleveref}
\usepackage{xcolor}
\usepackage{booktabs}
\newcommand{\summarybox}[2]{
\begin{quote}
{\it #1} #2
\end{quote}}
\usepackage{verbatim}
\usepackage{xspace}
\usepackage{subfig}
\usepackage{blindtext}

\newlength{\leftbarsep}
\colorlet{leftbarcolor}{black!120}

\usepackage{framed}
\renewenvironment{leftbar}{%
    \MakeFramed {\advance \hsize -\width \FrameRestore }%
}{%
    \endMakeFramed
}
\usepackage{listings}
\usepackage{textcomp}
\usepackage{xcolor}
\def\BibTeX{{\rm B\kern-.05em{\sc i\kern-.025em b}\kern-.08em
    T\kern-.1667em\lower.7ex\hbox{E}\kern-.125emX}}
    
\definecolor{codegreen}{rgb}{0,0.6,0}
\definecolor{codegray}{rgb}{0.5,0.5,0.5}
\definecolor{codepurple}{rgb}{0.58,0,0.82}
\definecolor{keywordsColor}{rgb}{0.000000, 0.000000, 0.635294}
\definecolor{backcolour}{rgb}{255,255,255}
\lstdefinestyle{mystyle}{
  backgroundcolor=\color{backcolour}, commentstyle=\color{codegreen},
  keywordstyle=\color{magenta},
  numberstyle=\tiny\color{codegray},
  stringstyle=\color{codepurple},
  basicstyle=\ttfamily\footnotesize,
  breakatwhitespace=false,         
  breaklines=true,                 
  captionpos=b,                    
  keepspaces=true,                 
  numbers=left,                    
  numbersep=5pt,                  
  showspaces=false,                
  showstringspaces=false,
  showtabs=false,                  
  tabsize=2,
  float=tp,
  floatplacement=tbp,
  xleftmargin=1.5em,
  belowskip=-8pt
}
\newcommand{\etal}[0]{\emph{et al.}\xspace}

\begin{document}

\title{Learning Defect Prediction from Unrealistic Data}

%\author{\IEEEauthorblockN{Anonymous Authors}
%}
\author{\IEEEauthorblockN{Kamel Alrashedy}
\IEEEauthorblockA{\textit{Georgia Institute of Technology} \\
\textit{School of Computer Science}\\
Atlanta, GA, USA \\
kalrashedy3@gatech.edu}
\and
\IEEEauthorblockN{Vincent J. Hellendoorn}
\IEEEauthorblockA{\textit{Carnegie Mellon University} \\
\textit{School of Computer Science}\\
Pittsburgh, PA, USA \\
vhellendoorn@cmu.edu}
\and
\IEEEauthorblockN{Alessandro Orso}
\IEEEauthorblockA{\textit{Georgia Institute of Technology} \\
\textit{School of Computer Science}\\
Atlanta, GA, USA \\
orso@cc.gatech.edu}
}

\maketitle

\begin{abstract}
Pretrained models of code, such as CodeBERT and CodeT5, have become popular choices for code understanding and generation tasks. Such models tend to be large and require commensurate volumes of training data, which are rarely available for downstream tasks. Instead, it has become popular to train models with far larger but less realistic datasets, such as functions with artificially injected bugs. Models trained on such data, however, tend to only perform well on similar data, while underperforming on real-world programs. In this paper, we conjecture that this discrepancy stems from the presence of distracting samples that steer the model away from the real-world task distribution.
To investigate this conjecture, we propose an approach for identifying the subsets of these large, yet ``unrealistic'' datasets that are most similar to examples in real-world datasets based on their learned representations. Our approach extracts high-dimensional embeddings of both real-world and artificial programs using a neural model and scores artificial samples based on their distance to the nearest real-world sample. We show that training on only the nearest, representationally most similar samples while discarding samples that are not at all similar in representations yields consistent improvements across two popular pretrained models of code on two code understanding tasks. Our results are promising, in that they show that training models on a representative subset of an unrealistic dataset can help us harness the power of large-scale synthetic data generation while preserving downstream task performance. Finally, we highlight the limitations of applying AI models for predicting vulnerabilities and bugs in real-world applications.

\end{abstract}

%\begin{IEEEkeywords}
%Pretrained models, fine-tuning, bug prediction, vulnerabilities prediction
%\end{IEEEkeywords}

\section{Introduction}

Applications of deep learning to defect\footnote{In the rest of the paper, we use the term ``defect'' in a somehow loose way to indicate both generic bugs and vulnerabilities.} prediction have gradually grown in number~\cite{Tantithamthavorn, Tantithamthavorn2, Jeongju}. This task poses a challenge for deep learning, as training larger, more performant models effectively requires large amounts of data \cite{kaplan2020scaling}. Datasets of real-world defects, however, typically contain at most a few thousands examples. While these datasets more than enough for other applications, such as debugging or program repair, they are inadequate for learning-based tasks.
This is part of a broader challenge in machine learning: collecting real-world data is often expensive, requiring significant human annotation effort, and resource-bound (e.g., there is a limited number of reproducible bugs in open-source projects).
Meeting the needs of ever larger, more data hungry models requires finding alternative, much larger sources of data.

Deep learning researchers have embraced two methods for improving model performance in the presence of limited real-world data: (1) separating the training of the model into multiple phases, and (2) synthetic data generation. The former separates model training into one or more \emph{pretraining} steps followed by a \emph{fine-tuning} step. While the fine-tuning consists of training the model on real-world data as it was done before, the pretraining steps initialize the weights of the model by training it on a larger corpus with a \emph{proxy} learning signal. Such signal may either be as generic as predicting randomly masked out tokens on billions of lines of code (as in BERT \cite{BERT_18}), or involve completing a task that mimics the downstream task, such as inserting vulnerabilities for a vulnerability prediction task. (Often, both approaches are applied in sequence, as we also do in our experiments.)

\begin{figure}[t!]
\centering
 {\includegraphics[width=0.60\linewidth]{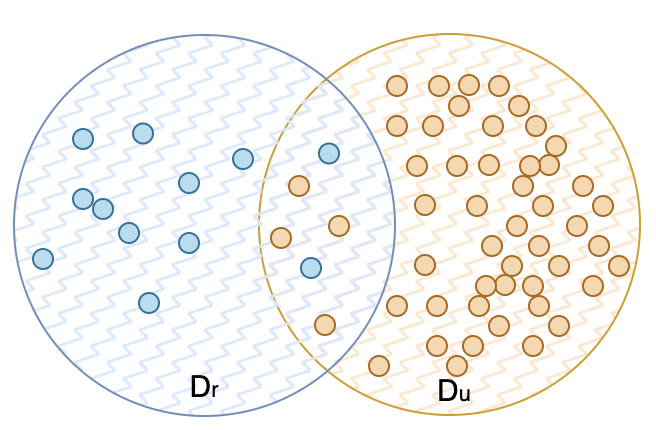}}
  \caption{Schematic depiction of the motivation for our work. Unrealistic datasets of bugs, $D_u$, are often far larger than datasets of real-world bugs, $D_r$, and contain many programs that are not representative of such real-world bugs.}
  \label{fig:venn}
\vspace{-8pt}
\end{figure}

This is where synthetic data generation comes into the picture, as it is often relatively easy to generate a large dataset with similar characteristics to downstream task data. In bug prediction, for instance, it has become common practice to artificially inject random bugs into millions of functions to train a model to detect and repair those bugs (e.g.,~\cite{vasic2019neural}).
Across machine learning applications, the use of synthetic data has in fact grown steeply, with some predicting that more than 60\% of the data used for training AI models will be synthetic by 2024~\cite{Alexander}.
Software engineering is no exception: synthetic and otherwise artificial datasets of defects abound (e.g.,~\cite{Draper,BUGLAB,hellendoorn2019global}).
The problem with such datasets is that they contain many examples that are not representative of real-world bugs (Fig \ref{fig:venn}), which can distract or even mislead the model during pretraining.  Consequently, \emph{deep learning models trained on synthetic data often perform poorly on real data}~\cite{BUGLAB,hellendoorn2019global}.

In this work, we attack this issue by proposing an approach for extracting a subset of ``most realistic data'' from a largely unrealistic dataset.
The goal of our approach is to preserve only those data that are likely to reinforce patterns that will benefit the model when used on real-world samples, while removing data that are highly dissimilar to any real-world ones. 
To do so, we leverage representation learning techniques, which are themselves based on deep neural networks, to map the available real-world data into a semantically rich, high-dimensional space.
We then identify the most realistic data by computing their similarity to representations of real-world data. Data that are far removed from any real-world data are unlikely to benefit (and may even negatively effect) the model during pretraining and are removed.

We empirically evaluate this approach on two downstream code understanding tasks---bug and vulnerability prediction---using two pretrained models of code across a range of hyper-parameter configurations. Our results are promising: our approach yields consistent improvements over both a traditional pretraining/fine-tuning setup and fine-tuning directly. We provide several relevant insights.
\textit{First}, the results simultaneously highlight that unrealistic dataset can contain (1) many samples that are representative of real-world ones, as well as (2) many distracting samples that should be removed.
\textit{Second}, they affirm recent findings that AI-powered systems can achieve higher performance by using \emph{less, but better} training data~\cite{sorscher2022beyond}. In several cases, in our evaluations, we find that using just the 10\% most realistic data in large, synthetic datasets can outperform any other configuration.
\textit{Finally}, they show that the way the unrealistic datasets are generated affects the ability of our approach to identify realistic data.

Overall, our evaluation shows that our approach can offer a balance between the ability to generate very large datasets offered by data generation and the need for such data to be faithful to the signal in the downstream task of interest, at least for the two tasks vulnerability and defect prediction.

%The rest of the paper is organized as follows. In Section \Cref{sec:motivation}, we discuss the motivation behind our research. The methodology of our research is discussed in \Cref{sec:Methodology}. We present how our approach is implemented and results in \Cref{sec:Empirical}. Our related works and conclusion sections are presented in \Cref{sec:Related} and \Cref{sec:conclusion}. 

\section{Motivation}
\label{sec:motivation}
We define \textit{unrealistic data} as datasets that are not necessarily representative of real-world code. Such datasets may contain samples that are similar in nature, or identical to, real-world programs, but typically contain a far larger number of implausible or incorrectly labeled programs. For instance, a defect can be synthetically introduced into a bug-free program by altering a single variable name. While this strategy \emph{could} introduce bugs that resemble real bugs found in public projects (as in Code Snippet \ref{real-world_code}, described later in this section), it is much more likely to yield unrealistic programs (as in Code Snippet \ref{synthetics_code}, also described later). Examples of synthetically created datasets include SATE IV Juliet \cite{SATE} and SARD \cite{SARD}. These datasets were originally generated to evaluate the performance of static analysis tools in vulnerability prediction~\cite{Saikat}. Another common type of unrealistic data are weakly labeled data, where a dataset is annotated using automated tools, such as static analyzers, that may suffer from a high rate of false positives and false negatives. For instance, the Draper dataset \cite{Draper} is comprised of potential bugs detected via static analysis in open-source software projects from GitHub, many of which are not true bugs.

Despite offering no strong guarantees of real-world efficacy, unrealistic datasets such as the ones we mentioned above are frequently relied upon in software engineering research because they address a core need of deep neural models: large volumes of data. Synthetic data generation mechanisms can in fact easily generate many millions of ``bug-like'' samples. 
The natural drawback of training on such data is that, for the reasons we provided above, many (when not most) samples within them poorly reflect real-world bugs. These synthetic bugs may not be problematic, or never occur, in practice, thus reinforcing patterns that reduce model performance on real-world bugs.
Such datasets are also often artificially balanced, with one or a few buggy samples paired to each bug-free sample (as in ETH Py150 \cite{ETH-Py}), or drastically overestimate the incidence of bugs by virtue of overly zealous rules (as in Draper).
Indeed, models trained on unrealistic data often perform \emph{very} poorly when tested on real-world bugs~\cite{Vuldeepecker, survey_Jie, Vulpecker}.
Perhaps as a consequence, most evaluations in this domain focus on unrealistic datasets with far higher rates of bugs than is typical in real-world settings.

\begin{lstlisting}[style=mystyle,escapechar=!,language=Python,label={synthetics_code}, caption={Example of an unrealistic, synthetically generated variable misuse bug.}]
# Buggy code
def __init__(self, parent=None, *args):
   QtGui.QListWidget.__init__(
      !\colorbox{light-gray}{args}!, parent, *args)
   self._take_names = []
   self.take_names = []

# Fixed code
def __init__(self, parent=None, *args):
   QtGui.QListWidget.__init__(
      !\colorbox{light-gray}{self}!, parent, *args)
   self._take_names = []
   self.take_names = []
\end{lstlisting}

%,basicstyle=\tiny
\begin{lstlisting}[style=mystyle,escapechar=!,language=Python,label={real-world_code}, caption={Example of a variable misuse bug found in real-world code.}]
# Buggy code
def __rel_change(self, new: float) -> float:
	if self._likelihoods:
		old = self._likelihoods[-1]
		return abs((new - old) / !\colorbox{light-gray}{old}!)
	return inf
	
# Fixed code
def __rel_change(self, new: float) -> float:
	if self._likelihoods:
		old = self._likelihoods[-1]
		return abs((new - old) / !\colorbox{light-gray}{new}!)
	return inf
\end{lstlisting}

The main challenge with using real-world datasets for learning defect prediction, in turn, is that they tend to be too small, often in the order of one thousand bugs. Furthermore, real-world defects tend to be non-trivial to detect, requiring broad contextual knowledge.
For instance, the real-world bug\footnote{From \url{https://github.com/yedivanseven/PLSA/commit/12eeb77e}.} shown in Code Snippet \ref{real-world_code} is far from obvious: a ``relative change'', as the method name indicates the function computes, could be measured both relative to the \lstinline{old} likelihood (buggy version, top) and the \lstinline{new} one (fixed version, bottom), depending on the intended logic. The method context alone does not make it clear why the former is incorrect.
In contrast, Code Snippet \ref{synthetics_code} shows an example of a synthetically induced ``variable misuse'' bug in a function (top) and its ``fixed'', original, version (bottom). This bug is unlikely to occur in practice, as the \lstinline{__init__} call already includes a reference to \lstinline{args}, and \lstinline{self} is always the first argument in such calls.

Table \ref{tab:dataset} shows the total number of samples of unrealistic data compared to the real-world ones for the two defect prediction tasks that we consider in this work. Consider, for instance, the bug prediction datasets (bottom rows). As the table shows, the ETH Py150 dataset is 75 times larger than the real-world one \cite{ETH-Py}. 
Allamanis et al. proposed a self-supervised approach to detect and repair the bugs in this dataset~\cite{BUGLAB}. They found that this model can detect and repair unrealistic bugs with high accuracy, yet yields substantially lower performance when tested on a second dataset, PyPIBugs, which contains a few thousand real-world bugs collected from code changes on GitHub. That study suggests that researchers should focus on improving the accuracy of the deep learning models on real-world bugs. In this work, we use PyPIBugs to show a new way to improve real-world performance in the presence of unrealistic data.

\begin{table}[t]
\centering
\caption{Total number of samples of unrealistic data compared to the real-world ones for the two defect prediction tasks that we consider in this work: vulnerability and bug prediction.}
\label{tab:dataset}
\begin{tabular}{c|cccccc}
\textbf{Task} & \textbf{Dataset} & \textbf{Realistic?} & \textbf{Samples} & \textbf{Annotated} \\
\toprule
\textbf{Vulnerability} & Draper  & No & 154K & Weakly labeled  \\
\textbf{prediction} & Devign & Yes &  27K &  Human labeled \\
\midrule
\textbf{Bug} & ETH Py150 & No & 306K & Synthetic  \\
\textbf{prediction} & PyPIBugs & Yes & 4.5K &  Human labeled  \\
\end{tabular}
\vspace{2mm}
\\
  % \kamel{ *It should be noted that PyPIBugs is the largest manually annotated real-world dataset for bug detection.}
\end{table}

\begin{figure*}[t]
  \centering
 {\includegraphics[width=.96\linewidth]{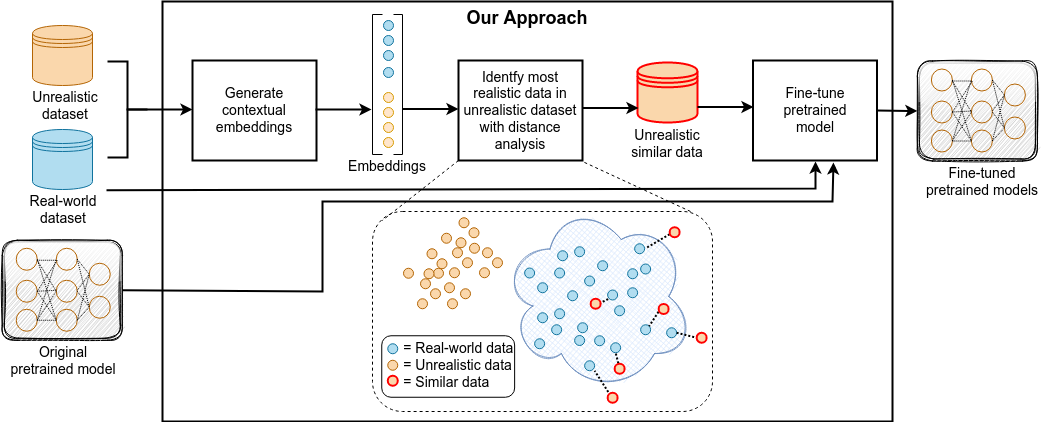}}
  \caption{Overview of our approach, which takes as input an unrealistic dataset, a real-world dataset, and a generically pretrained model, and produces as output a fine-tuned version of the pretrained model. To do so, it (1) converts the programs in the two datasets into a contextual embedding, (2) uses the distance between elements in the embedding to identify a subset of programs in the unrealistic dataset that are most similar to programs in the realistic one, and (3) uses the identified subset plus the realistic dataset to fine-tune the pretrained model.}
  \label{fig:schematic}
%  \vspace{-8pt}
\end{figure*}

\section{Our Approach}
\label{sec:Methodology}
 
The goal of this work is to improve the performance of models that are pretrained on large, unrealistic datasets for code understanding tasks, focusing specifically on bug and vulnerability prediction. We emphasize that we do not aim to achieve state-of-the-art results on these tasks, but rather to provide a \textit{general approach} for improving model performance in the presence of unrealistic datasets. 

The key intuition behind our approach is that it would be beneficial for these models to learn from the \emph{subset} of unrealistic data that are most representative of real-world ones. One of the main contributions of our approach, described in the rest of this section, is a technique for (1) detecting this subset and (2) leveraging it to improve model performance.

\subsection{Overview}

\Cref{fig:schematic} provides an overview of our approach, which takes as input (1) an unrealistic dataset, (2) a real-world dataset, and (3) a model that may have been pretrained on a large dataset (as is now common) and produces as output a fine-tuned version of the pretrained model. At a high level, our approach performs two main steps: (1) it converts the programs in the two datasets into a contextual embedding, so that all the programs are represented as vectors in the same space; (2) it applies vector similarity search technique to identify the subset of programs in the unrealistic dataset that are most similar to programs in the real-world one; and (3) it uses the identified subset plus the realistic dataset to fine-tune the pretrained model. In the rest of this section, we provide and clarify important terminology and describe these steps in detail.

\subsection{Background: Pretraining and Fine-Tuning}
Popular neural networks such as Transformers reliably improve with more training data and model capacity \cite{kaplan2020scaling}. However, datasets for most downstream tasks are limited in size---even datasets with hundreds of thousands of functions such as ETH Py150 \cite{ETH-Py} pale in comparison to the volume of code available on open-source platforms such as GitHub. Pretraining offers a solution: larger models are first trained on large volumes of unlabeled data to complete a fairly generic task, such as filling in a few masked-out tokens based on the context (BERT \cite{BERT_18}), or generating arbitrary code left-to-right (GPT \cite{gpt_3}), where they learn general representations of a broad spectrum of source code. Once initialized in this way, they can be fine-tuned on smaller datasets of downstream tasks, on which they then tend to perform substantially better than when trained on those from scratch. %The idea of pretrained models is similar to transfer learning, where the model is trained to learn one or more tasks, and then updated on a new, related task \cite{Xu_Han}.

Models initialized with such generic objectives typically learn general representation of many programs. However, they still struggle to learn a task as challenging as defect prediction from a small, real-world corpus, as we show in our results as well. Therefore, it is common practice to \emph{continue pretraining} such models on task-specific data, even if synthetic or otherwise unrealistic, so as to help the model learn task-relevant representations. Following this second pretraining phase, models can be fine-tuned on downstream task data. If the pretraining signal is sufficiently closely related to the data seen during fine-tuning, training will converge quickly, which in turn limits the risk of overfitting.
The second pretraining phase may also be described as a fine-tuning phase. To avoid ambiguity, we only use the term ``fine-tuning'' for the final training phase on downstream task data and mainly use ``pretraining'' to refer to training a model on (parts of) unrealistic data, regardless of the its original initialization.

\subsection{Generating Contextual Embeddings}
\label{sec:gener-cont-embedd}

As we mentioned above, our technique aims to improve the performance of models of code in the presence of (1) large, unrealistic and (2) smaller real-world datasets. More formally, given a dataset $\mathcal{D}_u$ of unrealistic programs, as defined in \Cref{sec:motivation} (e.g., weakly labeled or synthetically generated bugs) and a dataset $\mathcal{D}_r$ containing real-world bugs, we aim to identify a subset of programs $\mathcal{D}_{ru} \subset \mathcal{D}_u$ that maximizes model performance when first trained on $\mathcal{D}_{ru}$ for downstream task performance on $\mathcal{D}_r$.
We conjecture that we can approximate this optimum by identifying a subset  $\mathcal{D}_{\hat{ru}} \subset  \mathcal{D}_{u}$ of programs that are \emph{representationally} most similar to samples in $\mathcal{D}_{r}$, on the premise that such samples are among the most realistic ones in otherwise largely unrealistic data.

As a first step in identifying these samples, our approach converts all the programs from unrealistic and real-world into a contextual embedding, that is a vector representation in a shared space, using a well-calibrated neural network.

Word-level representations such as Word2Vec \cite{word2vec} and GloVec \cite{Glove} embed each token separately. This now classical family of techniques first converts a corpus of text to a series of tokens ($t_1$, $t_2$, ..., $t_n$) and then \emph{embeds} each token $t_i$ into a vector $\textbf{h}_i$. This embedding is globally consistent for each unique token---although a token may appear in different sentences and have multiple meanings, it is assigned the same vector representation at every occurrence. On the other hand, contextual embeddings convert each token $t_i$ into a vector based on its context ($..., t_{i-1}, t_i, t_{i+1}, ...$). That is, the same word may be assigned a different representation across occurrences. This allows contextual embeddings to reflect sequence-level, rather than token-level, semantics \cite{Qi_Liu}.

\subsubsection{Identifying the Most Realistic Data}

%\alex{TODO: Define ``Most Realistic Data'' upfront and use it consistently throughout instead of ``similar data'' and other terminology.}

After generating contextual embeddings, our approach uses vector similarity search to identify unrealistic examples with vector representations that are sufficiently ``near'' real-world ones: 
$\mathcal{D}_{\hat{ru}} = \{d_u \in \mathcal{D}_{u} | \min_{d_r \in \mathcal{D}_r} \delta(d_u, d_r) < k\}$ for a distance $k$. We refer to this as the \textbf{most-realistic data} in this work.
Later, we will show how we empirically estimate appropriate values for $k$ for different datasets.

Finding semantically similar samples, even if superficially distinct, is a common task in Machine Learning. For instance, in computer vision, two images may clearly depict the same bird even if their pixel-wise distance is very large. An example in our field is code search, where a neural engine takes a natural language description as input and finds code snippets that implement the same logic.
To find similar vectors, $k$ Nearest Neighbors (kNN) search based on the Euclidean distance between vectors is commonly used. Since neural models typically produce very high-dimensional vectors and the search space of neighbors may contain billions of examples, an Approximate Neighbor Search can reduce the computational cost of searching in such a large space. Researchers at Facebook (now Meta) introduced Facebook AI Similarity Search (FAISS), which uses various optimizations to speed up similarity search by some 8.5X \cite{FAISS}.

\begin{figure*}[t]
  \centering
  \subfloat[\textbf{Vulnerabilities Prediction:} Draper (weakly labeled) and Devign (Realistic)]{\includegraphics[width=0.49\linewidth]{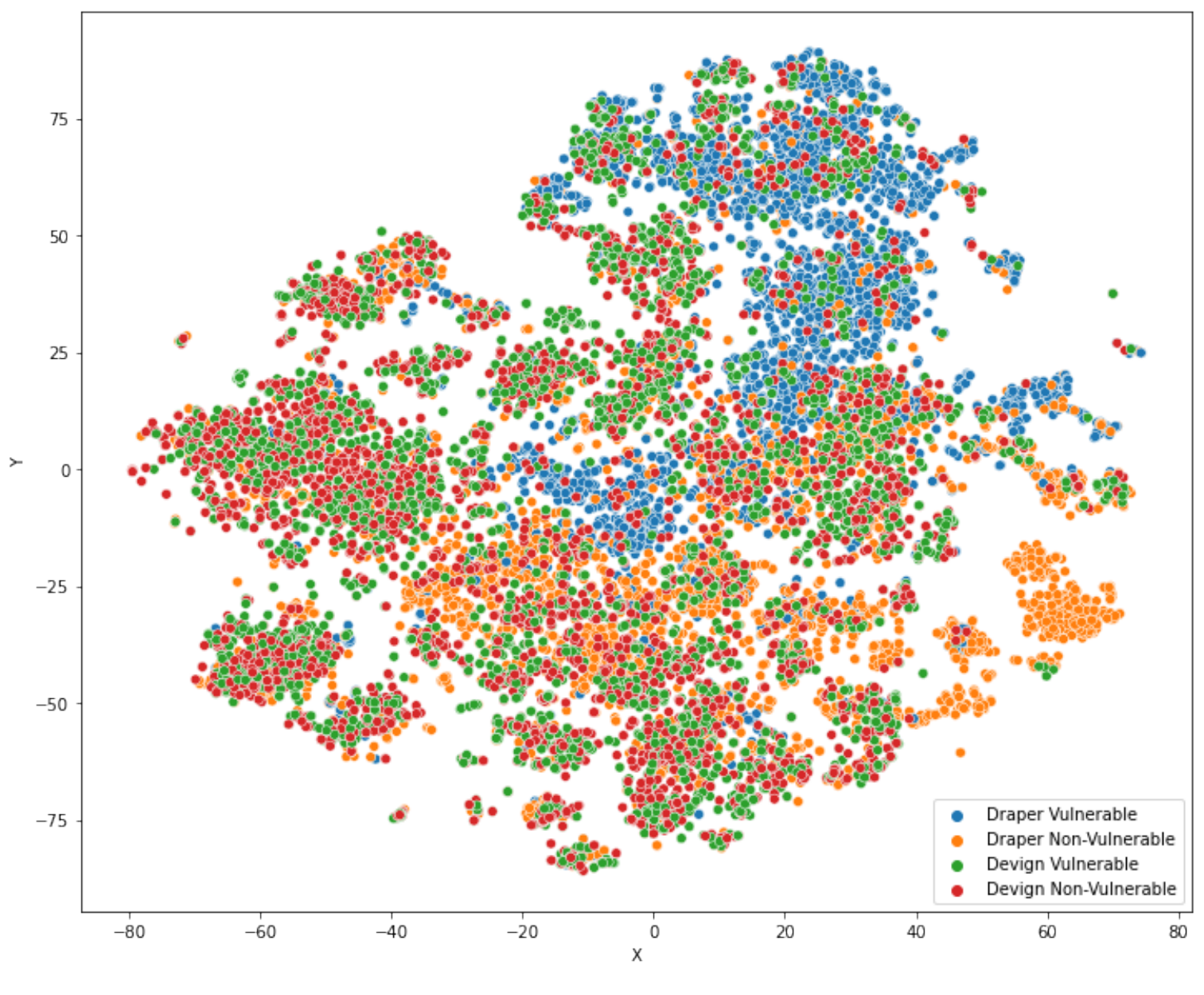}} % png
  \subfloat[\textbf{Bugs Prediction:} ETH Py150 (Unrealistic) and PyPIBugs (Realistic)]{\includegraphics[width=0.49\linewidth]{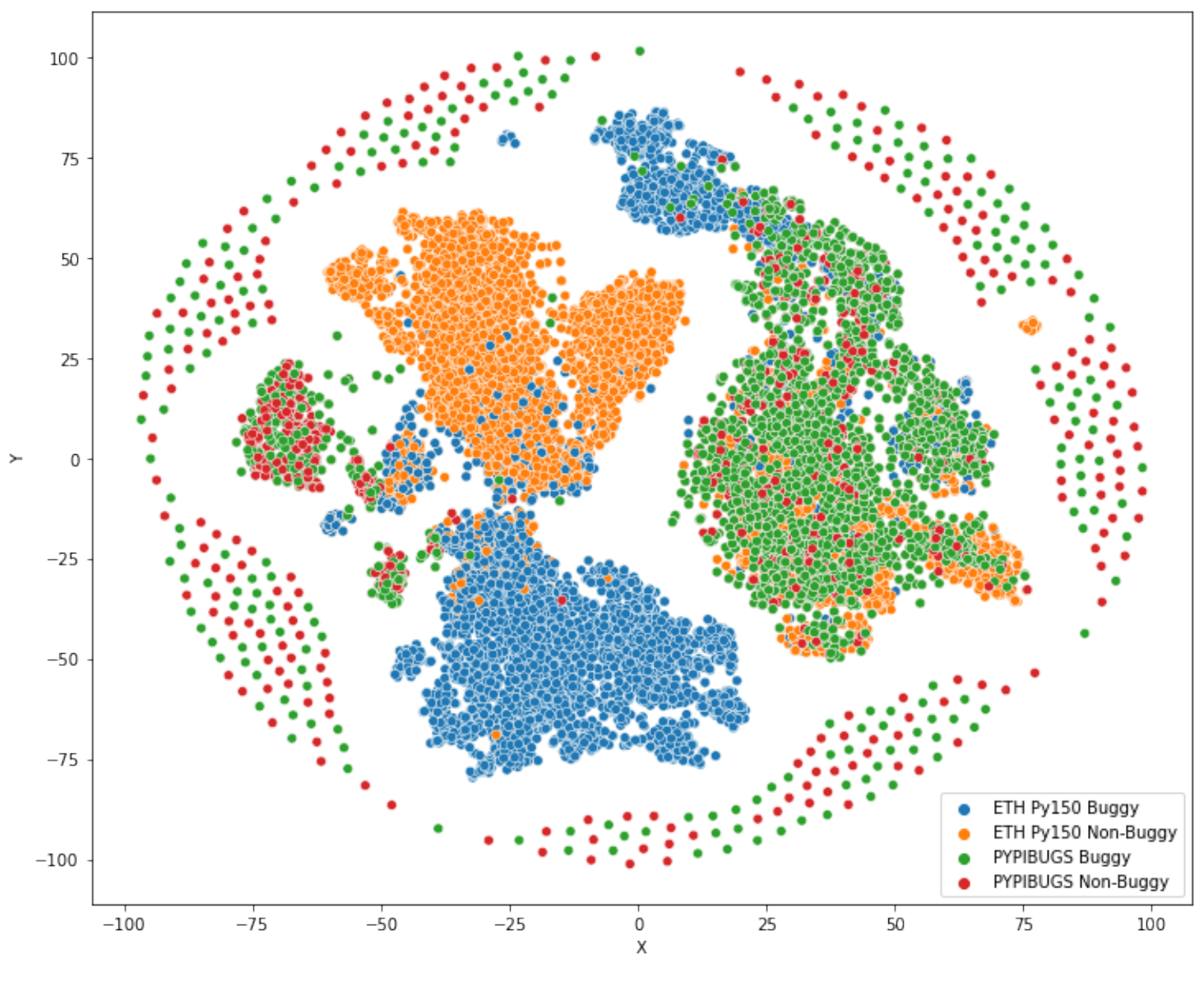}}%png
  \caption{High-dimensional embeddings of both real-world and unrealistic programs using T-SNE.}
  \label{fig:T-SNE}
\end{figure*}

\subsubsection{Fine-Tuning Pretrained Models}
\label{sec:training}

\begin{algorithm}[t]
\caption{The algorithm underpinning our approach. Given a (large) set of unrealistic programs and a (small) set of real-world ones, we identify a subset of unrealistic programs that are ``near'' real-world programs in a rich representational space. To extract such representations, we use a neural model, optionally fine-tuned with a generic objective on some of these programs to ensure that its representations accurately describe these programs.}
\label{alg:two}
\KwData{\\
// Two sets of programs $p \in P$ with labels $y \in Y$: \\
$\mathcal{D}_u := {p_u^0, ..., p_u^n},~ {y_u^0, ... y_u^n}$ \algorithmiccomment{Unrealistic data} \\
$\mathcal{D}_r := {p_r^0, ..., p_r^m}, ~{y_r^0, ..., y_r^m}$ \algorithmiccomment{Real-world data} \\
$\phi \in [0, 1]$ \algorithmiccomment{ Percentile of nearest points to include} \\
$\mathcal{M}: P \to \theta \in \mathbb{R}^k$  \algorithmiccomment{Model that embeds programs} \\
finetune $\in \{true, false\}$ \algorithmiccomment{ Whether to calibrate $\mathcal{M}$ on $P$}}

\KwResult{Realistic subset of unrealistic data, $\mathcal{D}_{\hat{ru}}$} 
\vspace{1em}
// Optionally, calibrate $\mathcal{M}$ on a subset of programs.

\If{finetune}{
    $\mathcal{M} = MLM\_train(\mathcal{M}, P_{random}), P_{random} \subset P_u$
}

// Embed all programs in $\mathcal{D}_u$,  $\mathcal{D}_r$

$\Theta_{\mathcal{D}_u} \in \mathbb{R}^{n \times k} = \mathcal{M}(p_u) ~\forall~ p_u \in P_u$

$\Theta_{\mathcal{D}_r} \in \mathbb{R}^{m \times k} = \mathcal{M}(p_r) ~\forall~ p_r \in P_r$

// Find Euclidean distance to nearest $\mathcal{D}_r$ for each $p_u$

$\Delta_{r,u} = \{\min_{\theta_r \in \Theta_{\mathcal{D}_r}} \delta(\theta_r, \theta_u)~\forall~\theta_u \in \Theta_{\mathcal{D}_u\}}$

// Set distance threshold at $\phi$th percentile of sorted distances.

$\vartheta$ = sorted($\Delta_{r,u}$)[$\lfloor \phi * |\mathcal{D}_u| \rfloor$]

//Return programs within distance $\vartheta$ of a real-world example.

$\mathcal{D}_{\hat{ru}} = \{p_u^i, y_u^i | \Delta_{r,u}^i \leq \vartheta\}$

\textbf{Return } $\mathcal{D}_{\hat{ru}}$
\vspace{1em}
\end{algorithm}

Our approach starts with generic pretrained models, which we calibrate using both the ``most realistic'' subsets of the unrealistic data, extracted using the method described in \Cref{alg:two}, and the (smaller set of) real-world data.
We use these datasets to pretrain and fine-tune the pre-trained models. Our main approach is a classic pretraining/fine-tuning setup, in which we first train the models on the most realistic data subsets and then fine-tune these models on real-world samples. We divide the extracted subset of unrealistic data into training and validation with ratios 98\% and 2\% respectively and continue training the pretrained models on the training portion with hyper-parameters spanning permutations of batch size \{16, 32, 64\} and learning rate \{3-e6, 1-e5, 2-e5, 3-e5\}. The model yielding the lowest validation loss is then fine-tuned on real-world data.

\section{Empirical evaluation}
\label{sec:Empirical}
In this section, we discuss how we implemented and evaluated our approach. We discuss the two tasks and four datasets (unrealistic and real-world) used in the empirical evaluation in \Cref{sec:tasks_and_data}. We then extract a range of ``most unrealistic'' subsets of the large datasets as described in \Cref{sec:extracting}. On each pair of datasets, we hyper-parameter tune, train and fine-tune two state-of-the-art models described in \Cref{sec:models}. Finally, we report the experimental results in \Cref{sec:results}.

\subsection{Tasks and Data Sets}
\label{sec:tasks_and_data}

Table \ref{tab:dataset} shows the two tasks and the corresponding four datasets we used in our evaluation. In this section, we discuss them in more detail.

\subsubsection{Vulnerability Prediction} 

This task consists of classifying functions as vulnerable or non-vulnerable. Vulnerability prediction is an important component of software security that helps protect software from attackers. In the past few years, applications of deep learning techniques to this task have become more popular (e.g., \cite{Saikat, Draper}). Vulnerability prediction is commonly included in evaluations of models of code \cite{CodeXGLUE}.

\textbf{Real-world Data:} The Devign dataset was collected from two open source repositories (QEMU and FFmpeg) containing ca. $26$K functions that were manually annotated by security teams \cite{Devign}. The initial data collection relied on identifying security-related issues in the messages associated with commits. Then, a security team manually annotated the functions as vulnerable or not. Several researchers have used Devign as a downstream task to evaluate their state-of-the-art pretrained models (e.g., \cite{CodeXGLUE,codet5,PLBART}).

\textbf{Unrealistic Data:} Draper is a large dataset, containing $154$K functions with potential vulnerabilities. It was collected from changes to public repositories on GitHub for which the commit message contained keywords such as ``buggy'', ``broken'', ``error'', or ``fixed''. Functions in this dataset were \emph{weakly labeled} through annotation using the static analysis tools Clang, Cppcheck, and Flawfinder \cite{Cppcheck,Clang,flaw}. Security researchers provided a mapping of findings from these tools to corresponding CWE types. For example, if Clang found that a function had an ``Out-of-bound array access'', it was labeled as vulnerable because this warning was related to ``CWE-805: Buffer Access with Incorrect Length Value''. Otherwise, the function was labeled as non-vulnerable. As these annotations were provided by static analysis tools, which are not always precise at detecting vulnerabilities, many examples in the Draper dataset are not representative of real-world vulnerabilities.

\subsubsection{Bug Prediction}

This task consists of detecting whether or not a function contains a bug. This type of dataset typically contains two or more versions of each function: the correct version and the one(s) containing a bug. Determining whether a function is buggy is cast as a binary classification task.
Deep learning approaches primarily target ``small'' bugs \cite{karampatsis2020often}, such as variable misuses \cite{vasic2019neural, hellendoorn2019global} and wrong operator bugs. Code snippets \ref{synthetics_code} and \ref{real-world_code} showed how real-world bugs of the first category can differ significantly in complexity from synthetic ones.

\textbf{Real-world Data:} The PyPIBugs dataset was collected by Allamanis \etal \cite{BUGLAB} by analyzing bugs reported in all PyPi Python packages\footnote{From \url{https://pypi.org/}}. PyPIBugs is the largest manually annotated real-world dataset for bug detection; however, the automatically collected data is poor, as shown in \cite{BUGLAB}. Due to the licensing limitations, the authors did not release the original data but provided supplementary material that allowed us to reconstruct the dataset. This dataset contains the buggy and fixed version of functions in real-world projects extracted from commits that indicate that a bug fix occurred and contained only a single, classifiable change to the source code. This is a small dataset, containing 2,289 buggy functions, each paired with its non-buggy counterpart for a total of 4,578 samples. PyPIBugs contains a variety of bug types, including variable misuse, swapped argument, and wrong binary operator detection.

\textbf{Unrealistic Data:} Lastly, ETH Py150 is a dataset of Python functions that has been used in several studies to generate a synthetic corpus of small bugs \cite{cubert, hellendoorn2019global}. Kanade \etal introduced five bug types to fine-tune their pretrained model, CuBERT \cite{cubert}, on bug classification. The authors applied the techniques recommended by Allamanis \cite{Allamanis_duplication} to remove duplicate code snippets using Jaccard similarity to avoid bias in the training.

\subsection{Extracting the Most Realistic Samples}
\label{sec:extracting}

As described in \Cref{sec:Methodology}, we aim to identify a subset of the large, yet unrealistic datasets that is most similar, in a learned representational space, to the bugs and vulnerabilities occurring in real development. The conjecture is that those samples are also the most likely to teach the model patterns of real-world defects.
We find such similar examples in two phases: (1) we convert the code snippets to vector embeddings using the OpenAI model \cite{cubert}, and (2) we search for similar vectors using the FAISS tool \cite{FAISS}. We discuss the details of this process here.

\subsubsection{Contextual Embedding}

We use OpenAI ``text-embedding-ada-002'' model to obtain a contextual embedding of source code functions \cite{cubert}. This model is a powerful embedding tool used to encode text/code snippets into embedding vectors. The OpenAI embedding model is utilized to find the most similar document and search for text or code. It can handle fairly long inputs, with up to 8,191 tokens. Its output is a single vector with a dimension of 1,536. %\kamel{Our findings indicate that the embedding provided by OpenAI are more suitable to this task than ones from CodeBERT, CuBERT, and CodeGen.}

In \Cref{fig:T-SNE}, we visualize the contextual embeddings of functions from both datasets: vulnerability prediction (Draper and Devign, left) and bug prediction (ETH Py150 and PyPIBugs, right) using a T-SNE plot---a low-dimensional (here, 2D) projection of high-dimensional representations that approximately preserves the pair-wise distances between data points \cite{t-SNE}. To avoid clutter, we randomly sub-sampled functions from the (larger) unrealistic datasets to match the number of functions in the (much smaller) real-world datasets. 

While the representations overlap in some areas, the plots show segments in which the unrealistic and real-world data are clearly separated. This effect is especially pronounced in the Python bug prediction data. Our conjecture is that such ``islands'' of unrealistic examples are unlikely to be useful when training a model to succeed on real-world samples and should be omitted from the training process. Note that the plots also show some isolated (groups of) real-world functions, which suggests that synthetic and weakly labeled data generation can fail to produce samples spanning the full space of true defects. This latter problem is not solved by our work but may be fruitful grounds for further improving models trained in this fashion (see \Cref{sec:discussion}).

At this point, we have contextual embeddings for all our data. Next, we discuss how to identify similar examples by using vector similarity search.

\begin{figure}[t]
    \centering
    \subfloat[\textbf{Vulnerability Prediction:} Distance distribution from 154K Draper (unrealistic) to their nearest Devign (real-world) samples.]{{\includegraphics[width=.9\linewidth, trim=0 0 0 10]{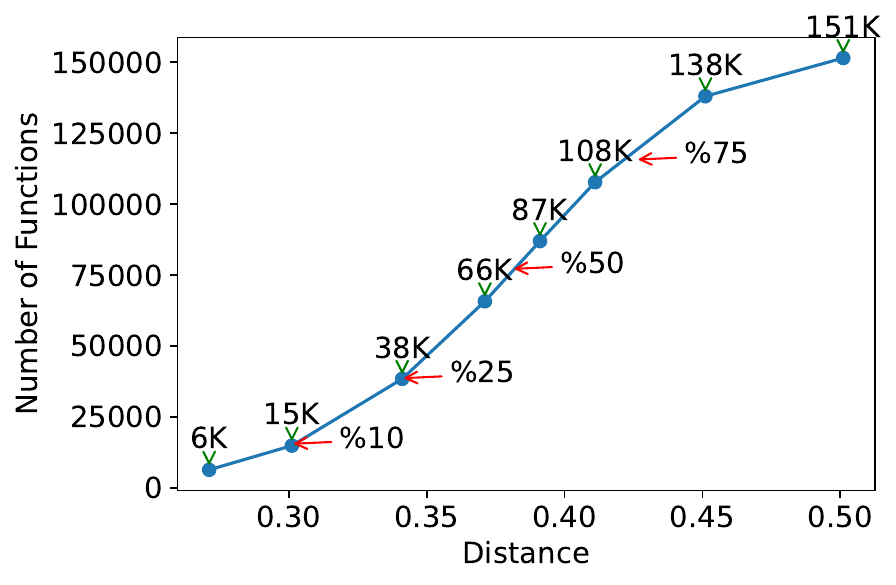} }} % Distance_Draper_to_Devign.eps
    \qquad
    \subfloat[\textbf{Bug Prediction:}  Distance distribution from 306K ETH Py150 (unrealistic) to their nearest PyPIBugs (real-world) samples.]{{\includegraphics[width=.9\linewidth, trim=0 0 0 10]{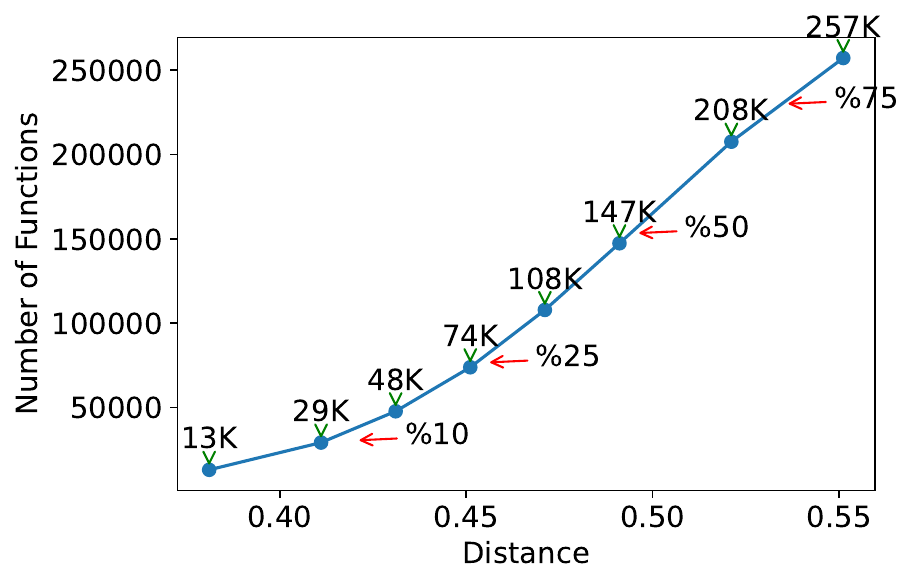} }} % Distance_ETH_py_to_PYPIBUGS.eps
    \caption{Euclidean distance (x-axis) and total number (y-axis) of functions of each set of unrealistic data to their nearest real-world data counterparts. We highlight the 10th, 25th, 50th and 75th percentile, for each of which we train models in our evaluation. Note that the most distant samples are not visualized, as these are up to four times as distant as those shown here.}
    \label{fig:Distance}
    \vspace{12pt}
\end{figure}

\begin{table}[t]
\centering
\caption{Total number of unrealistic samples and their distance to real-world ones for the two tasks considered.}
\label{tab:total_sample_distance}
\begin{tabular}{r|cccc}
\textbf{} & \multicolumn{2}{c|}{\textbf{Vulnerability}} & \multicolumn{2}{c}{\textbf{Bugs}} \\
\textbf{Tasks} & \multicolumn{2}{c|}{\textbf{ Prediction}} & \multicolumn{2}{c}{\textbf{ Prediction}} \\ \hline
\textbf{subset} & \textbf{Distance} & \textbf{Samples}  & \textbf{Distance} & \textbf{Samples}  \\
\toprule
%\textbf{5\%} & $\leq0.27$ & 7,707 & $\leq0.38$ & 15,336  \\
\textbf{10\%} & $\leq0.30$ & 15,528 & $\leq0.41$ & 30,515  \\
\textbf{25\%} & $\leq0.34$ & 38,676 & $\leq0.45$ &  76,363 \\
\textbf{50\%} & $\leq0.38$ & 78,186 & $\leq0.49$ &  153,622 \\
\textbf{75\%} & $\leq0.41$ & 115,985 & $\leq0.53$ &   230,105 \\
\midrule
\midrule
\textbf{100\%} & $\leq0.58$ & 154,150 & $\leq0.71$ & 306,729\\
\end{tabular}
\end{table}

\subsubsection{Similarity Search}

Following the extraction of contextual embeddings, we next index these representations into an efficient, approximate nearest neighbor search structure using the Facebook AI Similarity Search (FAISS) tool~\cite{FAISS}. We apply FAISS in our approach by building the data structure on vector embeddings of real-world data and then using the representations of unrealistic data points as queries. Each query returns the nearest neighbor and the distance to it, which we store to determine the ordering of all unrealistic samples in terms of their distance to the nearest real-world sample. The Euclidean distance between an unrealistic example's representation and that of its nearest real-world neighbor is calculated as: $\delta\left(\textbf{p},\textbf{q}\right) = \sqrt {\sum _{i=1}^{n} \left( \textbf{p}_{i} - \textbf{q}_{i}\right)}$.

The resulting distributions of distances for the two tasks we considered in our evaluation are visualized in \Cref{fig:Distance} and described in detail in \Cref{tab:total_sample_distance}. For vulnerability prediction, the farthest unrealistic sample was found at a distance of ca. 0.58 from its nearest real-world neighbor, but 75\% of the unrealistic data points were about four times closer (distances up to 615), and 10\% of the unrealistic data were again more than twice as close. A similar pattern held for the bug prediction datasets. In \Cref{sec:results}, we explore how these subsets can be used to improve the performance of pretrained models on real-world data.

Code Snippet's \ref{Draper_code} \& \ref{Devign_code} provide an example of the effectiveness of our technique. Code Snippet \ref{Draper_code} was extracted from the Draper dataset; Code Snippet \ref{Devign_code} is its nearest neighbor in the Devign dataset (distance: 0.13). The two share many similarities, including using the same parameter and return values, and both performing a character lookup based on addresses, just looking for different values respectively.

\subsection{Models and Training}
\label{sec:models}

Recent literature has proposed several pretrained language models of code that achieve strong results on code understanding and generation tasks~\cite{CodeXGLUE}.
We implemented our approach on top of four popular large language models of code. The first pretrained models (CodeBERT) are based on a Transformer encoder and are similar to the architecture of BERT~\cite{cubert,Codebert}.

\textbf{CodeBERT} \cite{Codebert} is a bimodal model pretrained on both source code and text with two objective tasks: (a) Masked Language Modeling and (b) Replaced Token Detection in seven different programming languages. Two different datasets were used to train CodeBERT: (a) an NL/PL dataset that includes 2.1M functions with comments, which helps the model learn the relationship between code and natural language; and (b) a set of 6.4M other functions to reinforce general code understanding abilities.

\textbf{CodeT5} \cite{codet5} is a pretrained encoder-decoder Transformer model introduced by Wang \etal. This model builds on the T5 ``Text-to-Text Transfer Transformer'' architecture \cite{t5}, which converts all language understanding and generation tasks into a shared text-to-text format, where each input is a text sequence prompt to be translated to a target phrase. The authors add two code-specific pretraining tasks: Identifier Tagging (IT) and Masked Identifier Prediction (MIP). CodeT5 was pretrained on CodeSearchNet data \cite{Codesearchnet} and additional C/C\# programs that were collected from GitHub repositories.

\subsection{Baseline}
Our approach aims to select a subset of a large, mostly unrealistic dataset that is representative of the real-world distribution. To evaluate the effectiveness of this technique, we compare our performance with three baseline alternatives.

\subsubsection{Fine-Tuning directly}
In this experiment, the pre-trained models were fine-tuned directly on a real-world dataset without using any unrealistic data (nicknamed 0\%, as it uses no unrealistic data). The results demonstrated how the pre-trained models perform when leveraging just a small, high-quality dataset.

\subsubsection{Fine-Tuning using the entire dataset}
In this experiment, we fine-tuned the pre-trained models using the entire unrealistic dataset (nicknamed 100\%). We compared the results of this experiment with those from our subset selection, which aims to remove noisy samples from this dataset.

\subsubsection{Random subset selection}
Finally, we compare the results of each subset to a baseline in which we train on a \emph{randomly} sampled subset of the unrealistic dataset of the same size. We contrast this performance with ours on a range of matching subset fractions, starting at 10\%. This comparison allows us to quantify the sample-effectiveness of our approach relative to a fair ground-truth value. It also occasionally provides a stronger baseline than both of the above. In a few experiments, mainly on vulnerability prediction, we found that training with the full unrealistic set substantially reduces performance, perhaps because its large size moves the pretrained model too far from its initial distribution, but using a small (e.g., 10\%) subset yielded slightly better performance than no fine-tuning at all.

\subsection{Experimental Results}
\label{sec:results}
\begin{table*}[t]
  \begin{center}
    \caption{Results of our approach on training models on (various most-realistic subsets of) unrealistic data and then fine-tuning them on real-world data, contrasted with fine-tuning without using any unrealistic data and random subset selection.}
    \label{table:fine-tuning_results}
    \begin{tabular}{l|l|l|cccccc}
      \toprule
       \textbf{Models} & \textbf{Approach}  & \textbf{Metric} & \textbf{100\%} & \textbf{75\%} & \textbf{50\%} & \textbf{25\%} & \textbf{10\%} &  \textbf{0\%}\\
      \midrule
      \multicolumn{8}{l}{\emph{\textbf{Vulnerability Prediction:} unrealistic data = Draper, real-world data = Devign}} \\
      \midrule

 CodeBERT & Baseline*   & Accuracy & 62.55  & 62.00 & 63.17 &  62.81 & 63.68   & 63.65\\ %~~(0.8\%, 1.8\%) \\
             &            & F1       & 53.09  & 54.71 & 55.40 &  55.12 & 55.31   & 54.42\\
             & Our selections    & Accuracy    & - & 63.54  & 62.81 & 64.23 & \textbf{64.53}   & -\\
             &                    & F1       & -  & 54.27 & 54.96 &  \textbf{62.29} & 59.37   & -\\

\midrule
     CodeT5 & Baseline*   & Accuracy & 61.53  & 62.92 & 62.81 &  62.70 & 63.39   & 63.17\\ %~~(0.8\%, 1.8\%) \\
             &            & F1       & 52.93  & 58.46 & 56.24 &  58.66 & 58.47   & 57.94\\
             & Our selections    & Accuracy    & - & \textbf{64.27}  & 63.61 & 63.57 & 63.20   & -\\
             &          & F1       & -  & 60.13 & 58.95 &  59.92 & \textbf{60.29}   & -\\
    
      \midrule
      \multicolumn{8}{l}{\emph{\textbf{Bug Prediction:} unrealistic data = ETH Py150, real-world data = PyPIBugs}} \\
      \midrule

   CodeBERT & Baseline*   & Accuracy & 58.51  & 57.20 & 58.51 &  55.90 & 57.42   & 53.05\\ %~~(0.8\%, 1.8\%) \\
             &            & F1       & 51.53  & 51.96 & 53.65 &  55.18 & 52.78   & 35.04\\
             & Our selections    & Accuracy    & - & 56.98  & 60.69 & \textbf{61.13} & 58.51   & -\\
             &                   & F1          & -  & 54.91 & 55.22 &  \textbf{60.26} & 56.01   & -\\

\midrule
     CodeT5 & Baseline*   & Accuracy & 59.38  & 57.42 & 58.51 &  57.64 & 55.67   & 50.0\\ %~~(0.8\%, 1.8\%) \\
             &            & F1       & 40.38  & 51.37 & 56.01 &  53.80 & 39.40   & 49.67\\
             & Our selections    & Accuracy    & - & 58.95  & \textbf{60.48} & 57.42 & 56.55   & -\\
             &                     & F1       & -  & 48.91 & \textbf{59.68} &  54.11 & 52.95   & -\\
    \bottomrule
    \end{tabular} \\
    \vspace{1mm}
    * The 0\% fine-tunes the model directly without employing any unrealistic data, while the 100\% fine-tunes it using the entire of unrealistic data. The 10\%, 25\%, and 75\% subsets are chosen randomly.

    % \vspace{-5mm}
  \end{center}
\end{table*}

In this section, we evaluate our approach to fine-tuning two pretrained models with two downstream tasks.

We continue training each of our models described above (in \Cref{sec:models}) on a range of subsets of the unrealistic data, sorted so that the smallest subset contains all unrealistic samples that are nearest to a real-world sample, and thus the most realistic according to our intuition. We conduct a hyper-parameter search over learning rates for each model, selecting the model that achieves the best validation loss within 10 epochs. Each model is then fine-tuned on real-world data in a second hyper-parameter sweep, again based on validation loss. The best-performing model is evaluated on the real-world test data. The resulting scores are presented in \Cref{table:fine-tuning_results}. 

For both downstream code understanding tasks (i.e., vulnerability and bug prediction), we evaluate two pretrained models in a variety of settings, ranging from direct fine-tuning on real-world data, pretraining on all (100\%) unrealistic data and then fine-tuning, and pretraining on various most-realistic subsets of unrealistic data.
Overall, our approach enhances the performance over the baseline settings (both fine-tuning directly, random subset selections and using the full pretraining data) on all tasks, sometimes by considerable margins. These gains required no modifications to the underlying model architectures, nor increases in training cost---in fact, models trained on just 10\% of unrealistic datasets often perform the best.

\subsubsection{Vulnerability Prediction}
We fine-tune CodeBERT and CodeT5 on a range of the ``most realistic'' subsets of the Draper dataset (i.e., 10\%, 25\%, 50\%, 75\%, and 100\% $\equiv$ all) and then fine-tune them on the Devign set. As a baseline, we fine-tune these pretrained models on randomly selected subsets (10\%, 25\%, 50\%, and 75\%) to compare the results with those obtained from our selected subsets. Each subset of Draper is randomly split between 98\% training and 2\% validation sets; Devign is already split into training, validation, and testing sets by \cite{CodeXGLUE}, so we use the same splits to ensure our results are comparable with those reported in prior work, shown in the left-most results column.

In general, the results show that the performance of the models improves over the baseline models when using subsets including 50\% of the total samples or less. Models trained on only samples that are closest to real-world data (i.e., the subset of 10\% and 25\%) mostly outperform models trained on larger volumes of data and randomly selected subsets (i.e.,
10\%, 25\%). This inversion of the correlation between data volume and model performance strongly supports our conjecture that the \emph{nearest samples are indeed more realistic}. 

Overall, the results with CodeBERT for the 25\% subset show an improvement of 1.42\% in accuracy (measured in absolute percentage points, not relative improvement) and of 7.17\% in F1 score compared to a randomly selected subset of 25\%. With our selected subset of 75\% on CodeT5, the accuracy improved by 1.35\% and F1 score by 1.67\% over the random subset of 75\% configuration. Additionally, there was a slight improvement over the randomly selected subset 75\% with increases of 1.35\% and 1.67\% in accuracy and F1 score, respectively.

In contrast, training the models on the entire Draper set and then fine-tuning them on Devign actually yields \ul{worse} performance than simply fine-tuning directly 
in our investigation -- CodeBERT's accuracy drops from 63.65\% to 62.55\% and its F1 score from 54.42\% to 53.09\%; CodeT5's accuracy drops from 63.17\% to 61.53 \% and its F1 score from 57.94\% to 52.93\%). This underscores our observation that many samples from Draper are poorly representative of real-world vulnerabilities and may indeed reduce the model's ability to detect such vulnerabilities.
%In the case of PLBART, removing even just the 25\% least-realistic outliers (going from 100\% and 75\%) yields the bulk of the increase in performance (1.02\% points).

\summarybox{\textbf{Finding \#1:} Vulnerability detectors pretrained on weakly labeled data can benefit from eliminating samples that are representationally far removed from real-world samples, improving  CodeBERT on the subset of 25\% by up to 1.42\% in accuracy and 7.17\% in F1 score on a Draper/Devign evaluation when trained on 4-10x less data.}

\subsubsection{Bug Prediction} 

We followed the same approach in vulnerability prediction, where we first pretrained the models on (a subset of) ETH Py150 and then fine-tuned them on PyPIBugs. We used two pretrained models (e.g., CodeBERT and CodeT5). %We also attempted to fine-tune PLBART on PyPIBugs with a range of hyper-parameter configurations, but consistently found that the training accuracy only exceeded 50\% (random guessing) by the time that overfitting had set in, so that the validation accuracy never improved beyond chance.

We are not aware of prior work that has fine-tuned these models on PyPIBugs, so as before, we produced baselines by both fine-tuning the models directly on PyPIBugs and by first fine-tuning on (subsets of) the ETH Py150 set and then fine-tuning on PyPIBugs. With the former, CodeBERT and CodeT5 performs is roughly on par with random guessing, with 53.05\% and 50.0\% test accuracy respectively (the datasets are balanced between buggy and bug-free programs). The F1 scores indicate particularly poor calibration on the part of CodeBERT (35.04). Pretraining on the entirety of ETH Py150 followed by fine-tuning somewhat improves accuracy scores to 58.51\% and 59.38\% on CodeBERT and CodeT5, respectively. This also improves CodeBERT's F1 score considerably, from 35.04\% to 51.53\%. Evidently, CodeT5 struggles more in this scenario; its F1 score drops to 40.38. We then compare our approach with these numbers, and with randomly selected subsets, on a range of pretraining dataset sizes. The results demonstrate that on 10\%, 25\%, and 50\% subsets, our approach outperforms the aforementioned baselines and almost always outperform the random selection baseline in both F1 score and accuracy. Conversely, the results for the 75\% subset indicates a drop in the accuracy of CodeBERT and the F1 score for CodeT5. Evidently, the 75\% subset contains a fairly large number of noisy and unrealistic samples that hurt the performance of the models.

%Interestingly, our subset of 10\% boosts CodeT5's F1 score by 13.55\%, compared to a random subset of the same size. This offers compelling evidence that our approach, which selects a small, realistic subset, helps to improve performance significantly better than a randomly selected subset that might contain noisy samples.

Our approach, meanwhile, shows significant improvement in the performance of the models when using between half and one-tenth as much unrealistic data, boosting the models' accuracy to 61.13\% and 60.48\%—an improvement of 5.23\% and 1.97\% resp., and the models’ F1 score to 60.26\% and 59.68\% —an improvement of 5.08\% and 3.67\%, respectively, over the equivalent random subsets. Here, the improvement over full pretraining is especially substantial, at ca. 2.62\% and 1.1\% points and F1 score of 8.73\% and 19.3\% points, respectively.

\summarybox{\textbf{Finding \#2:} Bug detectors pretrained on synthetically generated bugs can benefit from selecting for realism based on learned source code representations. Our approach reliably yields improvements in real-world bug prediction when training with between half and one-tenth of the synthetic samples. }

\section{Discussion}
\label{sec:discussion}

We proposed a new approach to improve the performance of pretrained models for application in real-world settings. In this section, we discuss several implications of our findings and promising avenues for extensions of this research. 

\paragraph*{A Richer View of Pretraining}

One way to interpret our approach is that we offer a \emph{continuous relaxation} of the traditionally discrete pretrain/fine-tuning paradigm. Fine-tuning directly amounts to using 0\% (none) of the pretraining data, and pretraining ordinarily involves using 100\% (all) such data. Our work offers a way to use values between these endpoints. To do this effectively, we need to identify the most useful subsets of the pretraining data. Usefulness here refers to the degree to which the selected samples enhance the performance on downstream tasks after both pretraining and fine-tuning. Given the nonlinear nature of the models involved, directly optimizing for this objective is currently infeasible. Based on our results, our approach for selecting such subsets is effective, but it is surely not ideal. For instance, it is possible that different sizes of subsets call for partially disjoint sets of examples, rather than the strict ordering that we enforce in the current definition of the approach.

We envision a line of research focused on finding algorithms that extract such subsets more effectively and for a wider range of tasks. Such methods should be evaluated based on both the cost of identifying the required subset and the Pareto front of gain versus the percentage of samples kept. For instance, a rivaling method might yield better results than ours with 50\% of the unrealistic data, but perform worse when using 10\% as much. Investigating and understanding this full spectrum of options and trade-offs would be informative for practitioners choosing between techniques. This task may be framed as balancing generalization and specialization. The models in our work are initially pretrained on general bodies of code. During training on synthetic datasets, they become specialists in the downstream task. However, since these datasets are not quite aligned with real-world datasets, the models can ``over-specialize" to the synthetic examples. The goal of this line of research is to find the right balance between generalization and specialization prior to fine-tuning.

\paragraph*{Synthetic Versus Weakly Labeled} 

As both of our T-SNE plots (\Cref{fig:T-SNE}) and vector distance analyses (\Cref{tab:total_sample_distance}) show, the way the unrealistic datasets are generated affects the performance of our approach. The samples in Draper (unrealistic, weakly labeled) are more likely to overlap with, and less likely to be far removed from, those in Devign (realistic) than the samples in ETH Py150 (unrealistic, synthetic) are to overlap with those in PiPYBugs (realistic). 
%A nontrivial number of PyPIBugs samples had \emph{no} nearby neighbor in ETH Py150, which is starkly evident in \Cref{fig:T-SNE}(b), and some synthetic bugs in ETH Py150 were far more distant to any real-world bugs in PyPIBugs (distances up to 0.71) than any Draper sample was relative to Devign (distances $\leq$ 0.58)
A nontrivial number of PyPIBugs samples had \emph{no} nearby neighbor in ETH Py150, as clearly shown in \Cref{fig:T-SNE}(b), and some synthetic bugs in ETH Py150 were far more distant to any real-world bugs in PyPIBugs (distances up to 0.71) than any Draper sample was relative to Devign (distances $\leq$ 0.58). In short, the weakly labeled data generation approach we considered tended to yield more realistic defects than its synthetic counterpart.
This result highlights new challenges: while our method can remove highly unrealistic outliers, its results are limited in realism by the underlying data generation technique. This indicates opportunities to create more accurate synthetic data generators for bug prediction. We believe that our metrics and techniques will be useful for guiding and evaluating such efforts.

\paragraph*{{Limitations}}
%\paragraph*{\underline{Binary classification in Software Engineering}} Applied AI in the Software Engineering domain for binary classification, limited to two tasks: vulnerability detection and bug detection. These tasks are very complicated and pose challenges in proposing AI models that achieve very high accuracy in balanced real-world dataset. Even human intelligence and domain experts (e.g., developers \& Software Security Engineers) are unable to distinguish between buggy and non-buggy functions. Our approach demonstrates a modest improvement in accuracy and a substantial enhancement in the F1 score for vulnerability detection. Notable improvement in the F1 score in bug detection.
%Our approach shows an improvement compared to the three baselines, but the results are still not robust enough for real-world applications. 
Our approach demonstrates an improvement in accuracy and F1-score compared to the three baselines. However, the scores we report are not yet viable for real-world applications. For instance, the accuracy on PyPIBug is around 60\%, and we note only a slight improvement in vulnerability detection.
This aligns with recent research findings that performance on real-world data is consistently low, for both vulnerability detection \cite{Saikat} and bug localization \cite{BUGLAB}.
Our work moves the field forward by showing that leveraging synthetic data alone can be made more effective with careful filtering. This allows us to improve performance substantially relative to a series of baselines. Given these consistent results on multiple models and datasets, our work should offer complementary benefits to improvements in models and training signals proposed in related work.

%Allamanis et al. \cite{BUGLAB} collected PyPIBugs, which is the largest real-world dataset manually annotated by human experts, and evaluated the entire dataset using a proposed model, BUGLAB, for self-supervised learning in bug detection and repair. They pointed out that detecting whether a snippet has a bug or not seems to be the hardest task, as no models have achieved higher than a 63\% accuracy. The dataset contains buggy functions and their fixed versions, highlighting the bugs made by developers in the real world. In Code Snippet \ref{real-world_code}, an example is provided in which the developer miscalculates the formula to compute the relative change, indicating variable misuse. Another example involves the Incorrect Comparison Operator, where developers improperly use the  \( < \), \( > \), and \( = \) symbols. These bugs present a significant challenge for developers to detect, and even fine-tuning CodeBERT and CodeT5 on bug detection achieve an accuracy of 53\% and 50\%, which is a random guessing. Our approach enhances performance to 60\% in both cases, which is significantly better than a random guess.}

\section{Related Work}
\label{sec:Related}

\underline{Deep learning for defect prediction:} Improving deep learning based defect prediction is a popular area of research at the intersection between artificial intelligence and software engineering \cite{hellendoorn2019global, BUGLAB, PLUR, Marc}. Allamanis \etal \cite{Allamanis_Graphs} and Hellendoorn \etal \cite{hellendoorn2019global} used synthetic bugs to train and evaluate bug detector models to detect variable-misuse bugs. DeepBugs \cite{DeepBugs} applied deep learning to detect three types of bugs: wrong operators, operands, and argument swappings. Dinella \cite{Hoppity} trained Hoppity on bugs extracted from commits on GitHub, an approach that can yield fairly large datasets of plausible, if not guaranteed, bug fixes.
Other approaches have focused on exploring new input formats and model architectures to improve performance. Chakraborty \etal convert programs to a graph representation called the Code Property Graph (CPG) and then train the Gated Graph Neural Networks to detect software vulnerabilities \cite{Saikat}. Fu \etal proposed VulRepair to automatically repair vulnerabilities based on T5 architecture in \cite{Michael}. Another approach uses a sequence-to-sequence model, called SequenceR, to fix buggy code \cite{Sequencer}. 
The contributions of our work are orthogonal to these: instead of training new defect detector models or creating new datasets, we propose a technique for extracting the most downstream-task-relevant subsets of pretraining datasets, which benefited all models we evaluated. 

\underline{Data Distribution Shifts:} Our work is not the first to consider the shift from unrealistic bugs to real-world bugs. Jingxuan \etal proposed a bug detector model that trains on synthetic bugs to learn the bug prediction domain and then trains the model on real-world bugs \cite{Martin}. During the training on real-world bugs, the authors applied a multi-task hierarchy, focal loss, and contrastive learning to improve performance. Ding \etal introduce another model that is pretrained on synthetic (unrealistic) bugs and then fine-tuned on real-world bugs to detect software vulnerabilities~\cite{Ding}, showing high performance on synthetic bugs that significantly dropped on real-world data. A jointly learned bug selector and bug detector was proposed by Allamanis \etal \cite{BUGLAB} with the aim to create real-world bugs for training the model and then testing the model on real-world bugs. The real-world bugs are not included in their training phase, however, so the model does not learn their characteristics directly. SemSeed learns from real-world bugs to create new realistic bugs that can be used for training bug detectors~\cite{SemSeed}. 
Our work is also complementary to these efforts. As discussed in \Cref{sec:discussion}, our approach is limited by the underlying data generation processes, so techniques that generate more realistic samples would benefit our approach. At the same time, our approach can reduce the burden of needing to generate \emph{only} realistic examples, as it can effectively remove many spuriously generated unrealistic examples. Outside of software engineering research, the machine learning community has created methods to remove noisy labels from training datasets \cite{chen2023learning, yao2023better}. Future work may study whether these techniques can be adopted to further improve performance on the tasks studied in this work.

\section{Conclusion}
\label{sec:conclusion}
Large, artificially generated datasets for downstream tasks usually contain many samples that are not representative of real-world data. In this work, we argue that it is possible to identify a subset of data within unrealistic datasets that are most similar to examples in real-world datasets based on their learned representations. To investigate our conjecture, we propose an approach and evaluate it on two defect prediction tasks. Our results show that training with even small numbers of the most realistic samples extracted using our technique consistently improves the real-world performance of multiple models across both considered tasks. These results highlight the potential for using less, higher quality data when training machine learning algorithms on source code, provide a path for exploring techniques that discover such data, and in general, motivate future research in this direction. We discuss several possible future work directions in \cref{sec:discussion}. Our data are available at \url{https://doi.org/10.5281/zenodo.10514652}

%\section*{Data availability}
%Our data are available at \url{https://doi.org/10.5281/zenodo.10514652}
%\balance

\section*{Acknowledgments}
\vspace{-4pt}
\begin{small}
  This work was partially supported by 
  % NSF Test migration
  NSF, under grant CCF-0725202,
  % DOE X-STACK Kokkos
  DOE, under contract DE-FOA-0002460,
  % gifts
  and gifts from Facebook, Google, IBM Research, and Microsoft Research.
\end{small}

\bibliographystyle{IEEEtran}
\bibliography{references}
\newpage
\section*{Appendix}

\begin{lstlisting}[style=mystyle, escapechar=!, language=Python, label={Draper_code}, caption={An unrealistic sample was extracted as a more realistic sample compared to the sample in Code Snippet 4, with a distance of 0.13.}]
aiff_probe(AVProbeData *p)
{
    if (p->buf[0] == 'F' && p->buf[1] == 'O' &&
        p->buf[2] == 'R' && p->buf[3] == 'M' &&
        p->buf[8] == 'A' && p->buf[9] == 'I' &&
        p->buf[10] == 'F' && (p->buf[11] == 'F' || p->buf[11] == 'C'))
        return AVPROBE_SCORE_MAX;
    else
        return 0;
}
\end{lstlisting}

%,basicstyle=\tiny
\begin{lstlisting}[style=mystyle,escapechar=!,language=Python,label={Devign_code}, caption={Example of in real-world code from Devign.}]
static int ape_probe(AVProbeData * p)
{
    if (p->buf[0] == 'M' && p->buf[1] == 'A' && p->buf[2] == 'C' && p->buf[3] == ' ')
        return AVPROBE_SCORE_MAX;
    return 0;
}

\end{lstlisting}

\end{document}